\documentclass{article}
\usepackage{PRIMEarxiv}
\usepackage[T1]{fontenc}
\usepackage{graphicx}
\usepackage{authblk}
\usepackage{multirow}
\usepackage{booktabs}
\usepackage{amssymb}
\usepackage{amsmath}
\usepackage{mathtools}  
\usepackage{amsfonts} 
\usepackage{cleveref}
\usepackage{subcaption}
\usepackage{pifont}
\usepackage{url}
\usepackage[UKenglish]{babel}
\usepackage{comment}
\usepackage{multirow}
\usepackage{wrapfig}
\usepackage{ragged2e}
\usepackage{xcolor}
\usepackage{paralist}
\usepackage{fancyhdr}

\fancypagestyle{firstpagefooter}{
  \fancyhf{} 
  \fancyfoot[L]{Preprint.} 
  \fancyfoot[C]{\thepage}
}
\begin{document}
\title{Training Data Reconstruction: Privacy due to Uncertainty?}
\author[1]{Christina Runkel}
\author[2]{Kanchana Vaishnavi Gandikota}
\author[3]{Jonas Geiping}
\author[1]{Carola-Bibiane Sch\"onlieb}
\author[2]{Michael Moeller}

\affil[1]{Department of Applied Mathematics and Theoretical Physics, University of Cambridge}
\affil[2]{Department of Electrical Engineering and Computer Science, University of Siegen}
\affil[3]{ELLIS Institute T\"ubingen, Max Planck Institute for Intelligent Systems, T\"ubingen AI Center}

\maketitle              
\begin{abstract}
Being able to reconstruct training data from the parameters of a neural network is a major privacy concern. Previous works have shown that reconstructing training data, under certain circumstances, is possible. In this work, we analyse such reconstructions empirically and propose a new formulation of the reconstruction as a solution to a bilevel optimisation problem. We demonstrate that our formulation as well as previous approaches highly depend on the initialisation of the training images $x$ to reconstruct. In particular, we show that a random initialisation of $x$ can lead to reconstructions that resemble valid training samples while not being part of the actual training dataset.
Thus, our experiments on affine and one-hidden layer networks suggest that when reconstructing natural images, yet an adversary cannot identify whether reconstructed images have indeed been part of the set of training samples.

\end{abstract}

\thispagestyle{firstpagefooter}
\section{Introduction}
The empirical success of modern deep learning heavily relies on training on large datasets which may potentially contain private and sensitive data. The trained models can memorise \cite{carlini2023quantifying} and replicate samples from training data \cite{somepalli2023diffusion}. Several recent works have focused on evaluating the vulnerabilities of deep networks in leaking information about training data using different attacks. In this work, we study the \textit{training data reconstruction} problem, i.e., trying to reconstruct (parts of) the data a network was trained on without additional knowledge about network gradients or the initialisation. 

Let us assume one has trained a neural network $\Phi(x_i, \theta^*)$ with training examples $(x_i,y_i) \in \mathbb{R}^{K} \times \mathbb{R}$, $i \in \{1, \hdots, m\}$, by minimizing the expectation of a loss function $\mathcal{L}$ over all training examples. Then we expect that $\theta^* = \theta(x,y)$ for 
\begin{equation}
\label{eq:lower_level_problem}
    \theta(x,y) \in \arg \min_{\theta} \frac{1}{m}\sum_{i=1}^m \mathcal{L}(\Phi(x_i; \theta),y_i).
\end{equation}
Therefore, the natural way to recover the training data $\{(x_i,y_i)\}$ a network $\Phi$ is trained on if the final weights $\theta^*$ are known, is to consider the bi-level optimisation problem 
\begin{equation}
    \label{eq:bilevel_problem}
        \min_{x,y} \; l(\theta^*, \theta(x,y)) \quad \text{s. t.} \quad \theta(x,y) \text{ solves \eqref{eq:lower_level_problem}}. 
\end{equation}

Interestingly, previous works in this area, most prominently \cite{haim2022reconstructing} and \cite{buzaglo2023deconstructing}, do not consider \eqref{eq:bilevel_problem}. Instead, the authors use an analysis of Karush-Kuhn-Tucker (KKT) conditions on certain equivalent margin maximisation problems to propose an approach which -- in our notation -- could be phrased as 
\begin{align}
\label{eq:gradient_penalty}
& \min_{x}\|\nabla_{\theta} E(x,y; \theta^*) \|_2^2\\
    \text{ for } & E(x,y;\theta) = \frac{1}{m}\sum_{i=1}^m \mathcal{L}(\Phi(x_i; \theta),y_i).
\end{align}
As shown in \cite{Geiping_2019_ICCV}, the above can be interpreted as an approximation to the original cost function. More specifically, \eqref{eq:gradient_penalty} becomes a majoriser of \eqref{eq:bilevel_problem} (and can be used in iterative algorithms) if the costs $E$ are strongly convex and the upper-level loss $l$ is strongly smooth. 

While \cite{haim2022reconstructing,buzaglo2023deconstructing} demonstrated quite remarkable reconstructions of training data using \eqref{eq:gradient_penalty} in image classification, we focus on the robustness and the faithfulness of such reconstruction with respect to different initialisations for both \eqref{eq:bilevel_problem} and \eqref{eq:gradient_penalty}. Our numerical experiments demonstrate that while a carefully chosen initialisation recovers some examples $x_i$ from the training data, there also exist cases where somewhat realistically looking data samples $x_i$ are recovered that are \textit{not} part of the training data: The energy landscapes of both \eqref{eq:bilevel_problem} and \eqref{eq:gradient_penalty}, seem to have many minima, and initializing either approach with realistically looking images that have not been part of the training data quickly leads to a (numerical) convergence without resembling the training data at all. Thus, we argue that -- despite the ability to reconstruct some training examples -- the uncertainty of whether realistically looking images have been part of the training data preserves some privacy.

\section{Related Work}
Several prior works have studied different privacy vulnerabilities of deep networks. Membership inference attacks \cite{shokri2017membership,nasr2019comprehensive} ascertain whether a sample belongs to the model's training set. Attacks to reconstruct training samples using gradient updates \cite{phong2017privacy,zhu2019deep,zhao2020idlg,geiping2020inverting,zhu2020r} are relevant in the context of federated learning, where a global model is trained at a centralised server via gradient updates shared by the collaborating users. Such attacks are possible despite averaging gradients over large batches \cite{yin2021see}, are extended to newer architectures like vision transformers \cite{hatamizadeh2022gradvit}, and have been further improved using generative priors \cite{jeon2021gradient}. Recent work \cite{zhang2023understanding} even proposed a defense against such gradient inversion attacks. An alternative line of work considers the problem of recovering a single training data point \cite{balle2022reconstructing,guo2022bounding,ye2023initialisation} from a network trained using differentially private training assuming an informed adversary with access to model weights and all the remaining training samples,  and \cite{hayes2023bounding} additionally assume knowledge of intermediate gradients.
All the aforementioned attacks consider that the adversary has information in the form of samples or gradient updates that may not be available. Model inversion attacks \cite{fredrikson2015model,wang2021variational} attempt to recover samples representative of the classes in the original training set with access to only the trained network, yet these may not be the samples of the actual training set.

 In contrast to these works, Haim et al.~\cite{haim2022reconstructing} propose a privacy attack that recovers a subset of samples from training set using only the parameters of a trained neural network. They utilise the results in \cite{Lyu2020Gradient,ji2020}  showing convergence in the direction of margin maximization for homogenous neural networks trained with gradient descent for binary classification, and propose an attack to recover the samples on the margin.  The success of this attack depends on several assumptions -- it needs a carefully designed initialisation scheme where the variance of initial weights in the first layer is small,  the network should be without bias terms,  and is trained until convergence to reach a KKT point. This was extended in \cite{buzaglo2023deconstructing} to  multi-class classification and general loss functions. 

 Loo et al.~\cite{loo2023dataset} propose a dataset reconstruction attack that can recover the whole training set with the knowledge of model weights and initialisation in the infinite width regime when networks are trained using mean squared error loss. They show that the empirical success of this attack depends on deviation from the infinite width limit, and that the samples recovered from such an attack can be used as a distilled dataset \cite{wang2018dataset} for training networks. In our work, we formulate training data reconstruction as a bi-level problem considering the knowledge of only the trained model weights similar to \cite{haim2022reconstructing,buzaglo2023deconstructing}.

\section{Reconstructing Training Data}

\subsection{The linear case}
Consider a linear network architecture for predicting $N$ many labels $y_i \in \mathbb{R}^L$ from $N$ many corresponding input data points $x_i \in \mathbb{R}^K$, via an affine linear network architecture. For the sake of simplicity, we denote by $\bar{X} \in \mathbb{R}^{N\times (K+1)}$ the matrix formed by taking all $x_i$ with an additionally appended $1$ as rows, and $Y \in \mathbb{R}^{N\times L}$ the matrix obtained from using $y_i$ as rows. Then the affine linear network architecture $\phi(x_i;\theta) = \theta (x_i,1)^T $ leads to the training problem
\begin{equation}
\label{eq:linearCase}
    \min_\theta \mathcal{L}(\bar{X}\theta, Y)
\end{equation}
(where the above notation slightly abuses/extends our definition of a loss function to a matrix-values version and refers the the averaging of the individual losses along the batch-dimension). The optimality condition yields
\begin{equation}
\label{eq:linearOptimality}
    0 = \bar{X}^T \nabla \mathcal{L}(\bar{X}\theta^*, Y),
\end{equation}
where $\nabla \mathcal{L}$ refers to the gradient in the first component of the loss. In the case of training data reconstruction, we may assume that someone gives us the network architecture, loss function, and optimal parameters $\theta^*$ that satisfy \eqref{eq:linearOptimality}, and our goal is to reconstruct $\bar{X}$.

For a majority of loss functions, we may assume that $\nabla\mathcal{L}(\bar{X}\theta^*, Y)=0$ if $\bar{X}\theta^*= Y$ as we would have achieved a perfect fit of our desired outputs. While $\bar{X}\theta^*= Y$ is often impossible when optimising for $\theta$, it is extremely easy if both $\bar{X}$ and $Y$ are unknowns to optimise for, with $Y = 0$, $\bar{X} =0$ being a trivial (but unreasonable) solution. Even if we assume $Y$ to be known (e.g. by knowing there are equally many training examples for each class of a binary classification problem), 
the sufficient optimality condition $\bar{X}\theta^*= Y$ induces only $L\cdot N$ (number of training examples times dimension of the desired output) linear equations for $N\cdot K$ (number of training examples times dimension of the input data) many equations. For problems like the binary classification of images (as investigated in \cite{haim2022reconstructing}), $K>>L$, such that this problem is extremely underdetermined already. In addition, we do not need to satisfy  $\bar{X}\theta^*= Y$ - it would be sufficient if $\nabla \mathcal{L}(\bar{X}\theta^*, Y) \in \text{ker}(\bar{X}^T)$, which further enlarges the possible solution space. 
\subsection{The general case}
The general case (with arbitrary network architectures) is more difficult to analyse. Yet, two principle observations are worth noting. First, the more parameters a network has, the more equations the optimality condition yields, as each partial derivative w.r.t. each weight has to be zero. Thus, at the point where one has at least $N\cdot K$ (number of training examples times dimension of the input data)  many parameters and may still assume to know the desired output $Y$, one has significantly better chances to recover the training data. Yet, the optimality conditions are not linear and typically strongly coupled. In particular, according to the chain rule, the gradient of every parameter will contain a factor $\nabla \mathcal{L}(\phi(\bar{X};\theta), Y)$. Thus, it still holds that if there are input data $\bar{X}$ that allows the network to fit the desired output exactly, all optimality conditions are automatically satisfied. In particular, for classification where many rows of $Y$ are identical, it is sufficient to find \textit{a single example} $x_i$ that leads to the desired output and merely repeat this example many times in an identical manner\footnote{We note that a typical architecture for classification involves a softmax function, which by construction cannot satisfy $\phi(\bar{X};\theta)=Y$ for binary labels (and the training problem is not well-defined without additional regularisation such as a weight decay). Yet, network outputs close to $y_i$ numerically have the same effect as equality.}, such that the training data reconstruction can easily collapse. Moreover, this collapse can happen with any example that the final network classifies correctly (with extremely high confidence, i.e., with a network output very close to  $1$). 

Therefore, the ability to reconstruct training data has to depend on the luck (or clever initialisation and regularisation) of converging to a good local minimum with diverse and realistic examples $x_i$. We will showcase in Sec.~\ref{sec:experiments} that this can partly be achieved, yet, the abovementioned collapse frequently occurs and  seemingly realistic data cannot be guaranteed to be part of the training set.

\section{Numerical experiments}\label{sec:experiments}
\begin{table}[b]
    \centering
    \caption{Results highlighting the average distance of ground truth parameters $\theta^*$ and parameters $\theta$ after solving the bilevel optimisation problem for  affine and one-hidden layer classifiers per experiment for a random initialisation of $\theta$.}
    \label{tab:affine_theta}
    \begin{tabular}{c  |c | c} \hline
          \multirow{2}{*}{$x$\textsubscript{initialisation}} & Affine&One-hidden\\\cline{2-3}
           &~~$\|\theta^* - \theta\|_2^2$~~&~~$\|\theta^* - \theta\|_2^2$~~\\ \hline
         Random &~~$1.1214 \cdot 10^{-3}$~~&~~$6.1157 \cdot 10^{-5}$~~\\
         Ground truth~~&~~ $1.3888 \cdot 10^{-3}$~~ &~~$1.7055 \cdot 10^{-3}$~~\\ 
         CIFAR partition &~~$2.9749 \cdot 10^{-3}$~~ &~~$6.3629 \cdot 10^{-3}$~~\\ 
 \hline
    \end{tabular}
\end{table}
\begin{figure*}[t]
\centering
\begin{subfigure}[b]{0.48\textwidth}
    \includegraphics[trim=0cm 2cm 9.66cm 0cm, clip, width=\textwidth]{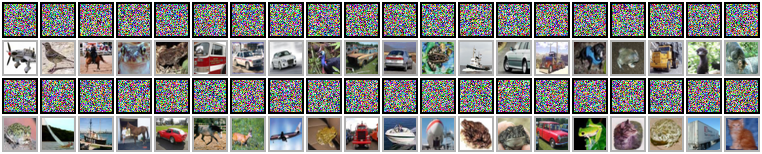}
\caption{Affine classifier.}
\label{fig:linear_classifier_1b}
\end{subfigure}
\begin{subfigure}[b]{0.48\textwidth}
    \centering
    \includegraphics[trim=0cm 2cm 9.66cm 0cm, clip,width=\textwidth]{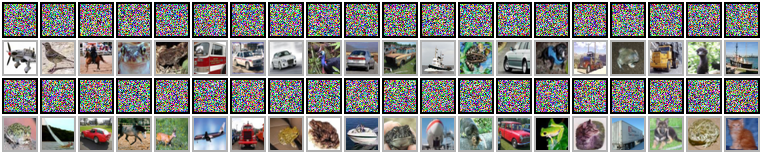}
    \caption{One-hidden layer classifier.}
    \label{fig:one_hidden_layer_1b}
\end{subfigure}
\caption{Reconstructions of training samples (black bounding box (BB)) and their nearest neighbour of the dataset (gray BB) for a random init. of $x$.}
\label{fig:rand_theta_rand_x}
\end{figure*}
\subsection{Influence of the initialisation}
For many non-convex optimisation problems, it is infeasible to compute global minimisers. In particular, problem \eqref{eq:bilevel_problem} in general cannot be solved exactly and one has to approximate the minimizer $\theta(x)$ of the lower-level problem by running a suitable descent scheme on $E$ (starting from some $\theta^0$) and initialising the upper-level problem by a suitable initial guess $x^0$. 
In the following, we analyse the influence of different initialisations for the proposed bilevel optimisation formulation as well as the approach of Buzaglo et al.~\cite{buzaglo2023deconstructing} and demonstrate that initialisation can be  crucial for obtaining good results in both approaches.

\subsection{Numerical Implementation}
We compare a stochastic gradient descent with momentum implementation of \eqref{eq:gradient_penalty} with a Quasi-Newton method for solving the bilevel problem~\eqref{eq:bilevel_problem} with \textit{implicit differentiation}, see e.g. \cite{lorraine2020optim}. The resulting update steps then read as
\begin{equation}
    \begin{cases}
        \theta^{k+1} &= \arg \min_{\theta} E(x^k, y; \theta), \\
        p^{k+1} &= H^{k+1} \big(\nabla l(\theta^*, \theta^{k+1}) \big), \\
        x^{k+1} &= x^k - \eta \nabla a(x^k).
    \end{cases}
\end{equation}
for a learning rate $\eta \in \mathbb{R}_{+}$, $a(x^k)= \nabla_x \langle \nabla_{\theta} E(x^k, y; \theta^{k+1}), p^{k+1}\rangle$ and $H^{k+1} $ being a Quasi-Newton approximation of $\big[\nabla_{\theta}^2 E(x^k, y; \theta^{k+1})\big]^{-1}$.
For this, we need the energy $E$ to be twice continuously differentiable, with invertible Hessian at all $\theta^{k+1}$. Further, $x^k$ and $\nabla_{\theta}E(x^k, y; \theta^{k+1})$ must be differentiable with respect to $x$.

\begin{figure*}[t]
    \centering
    \begin{subfigure}[b]{0.48\textwidth}
        \includegraphics[trim=0cm 2cm 9.66cm 0cm, clip,width=\textwidth]{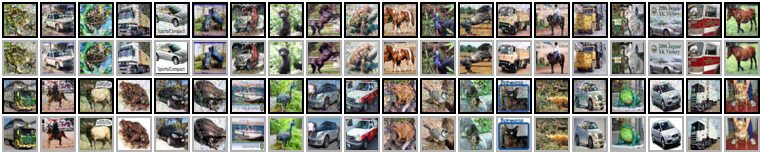}
        \caption{Affine classifier.}
        \label{fig:linear_classifier_2c}
    \end{subfigure}
    \begin{subfigure}[b]{0.48\textwidth}
            \includegraphics[trim=0cm 2cm 9.66cm 0cm, clip,width=\textwidth]{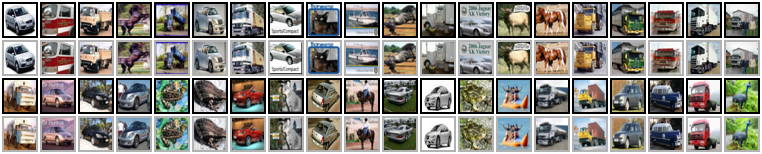}
        \caption{One-hidden layer classifier.}
        \label{fig:one_hidden_layer_2c}
    \end{subfigure}
\caption{Reconstructions of training samples (black BB) together with their nearest neighbour of the dataset (gray BB) for a ground truth init. of $x$.}
\label{fig:rand_theta_GT_x}
\end{figure*}
\begin{figure*}[t]
    \centering
\begin{subfigure}[b]{0.48\textwidth}
    \includegraphics[trim=0cm 2cm 9.66cm 0cm, clip,width=\textwidth]{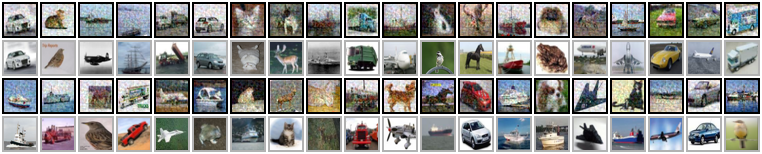}
    \caption{Reconstructions (black BB) vs nearest neighbour of training set (gray BB).}
    \label{fig:linear_classifier_3b}
\end{subfigure}
\hfill
\begin{subfigure}[b]{0.48\textwidth}
    \includegraphics[trim=0cm 2cm 9.66cm 0cm, clip,width=\textwidth]{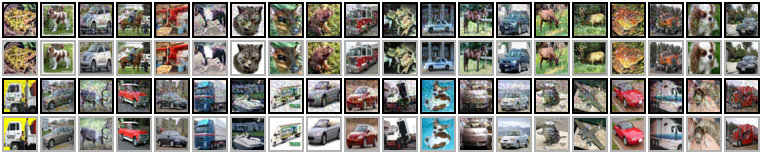}
    \caption{Reconstructions (black BB) vs nearest neighbours of CIFAR partition used for init. $x$ (gray BB).}    \label{fig:linear_classifier_3b_CIFAR_init}
\end{subfigure}
\caption{Reconstruction of training samples for \textit{affine classifier}, CIFAR partition init. for $x$.}
\label{fig:linear_classifier_3}
\end{figure*}
\begin{figure*}[t]
    \centering
\begin{subfigure}[b]{0.48\textwidth}
        \includegraphics[trim=0cm 2cm 9.66cm 0cm, clip,width=\textwidth]{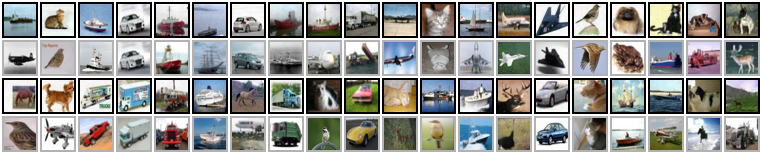}
        \caption{Reconstructions (black BB) vs nearest neighbour of training set (gray BB).}
    \label{fig:one_hidden_layer_3b}
    \end{subfigure}
    \hfill
    \begin{subfigure}[b]{0.48\textwidth}
    \centering
    \includegraphics[trim=0cm 2cm 9.66cm 0cm, clip,width=\textwidth]{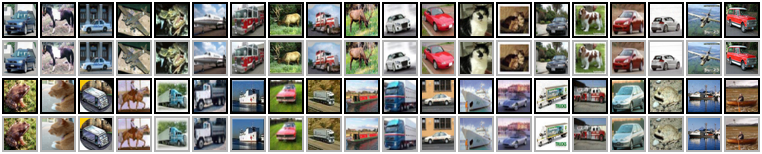}
    \caption{Reconstructions (black BB) vs nearest neighbours of CIFAR partition used for intialising $x$ (gray BB).}
    \label{fig:one_hidden_layer_3b_CIFAR_init}
    \end{subfigure}
\caption{Reconstruction of training samples for \textit{one-hidden layer classifier}, CIFAR partition init. for $x$.}
\label{fig:one_hidden_classifier_3}
\end{figure*}
\subsection{Bilevel optimisation problem} \label{subsec:bilevel_optimisation_problem}
We first analyse the reconstruction of images from an affine classifier
\begin{equation}
    \Phi(x; \theta) = x\theta^T_w + \theta_b
\end{equation}
for a weight matrix \(\theta_w \in \mathbb{R}^{3072 \times 1}\) and a bias term \(\theta_b \in \mathbb{R}\).
In a second step, with the aim to show that the results of the affine classifier generalise to other neural network architectures, we train a one-hidden layer neural network 
\begin{equation}
    \Phi(x; \theta) = \Big(\sigma(x\theta_{w_1}^T+\theta_{b_1})\Big)\theta_{w_2}^T+\theta_{b_2}\
\end{equation} 
for a ReLU activation function $\sigma$, weigths $\theta_{w_1} \in \mathbb{R}^{3072 \times 6144}$ and $\theta_{w_2} \in \mathbb{R}^{6144 \times 1}$ and bias terms $\theta_{b_1}$ and $\theta_{b_2}$. We use a full batch training setting for the classifier and solve a binary classification problem on the CIFAR10 dataset \cite{krizhevsky2009learning}.
For simplicity, in the following we will denote the set of all parameters as $\theta$.

\noindent To analyse the effect of different initialisations, we analyse the following settings:
\begin{enumerate}
    \item[a.] \textbf{Random initialisation of $x$}: Testing what would be the biggest breach in privacy, we randomly initialise the input $x$ by drawing values from $\mathcal{U}(0, 1)$. 
    \item[b.] \textbf{Ground truth initialisation of $x$}: As a proof of concept, we initialise the reconstruction algorithm with ground truth training images $x$. 
    \item[c.] \textbf{Initialisation of $x$ with a different partition of the dataset}:
    We divide the dataset into two non-overlapping subsets and use samples from one partition to train the classifier and initialize the reconstructions from the other partition.
    We analyse if the reconstruction method returns the ground truth training images or a different reconstructions. 
\end{enumerate}
In the following, we examine the influence of the different types of initialisations of $x$ while initialising $\theta$ by drawing every weight independently from a standard Uniform distribution for an affine and one-hidden layer classifier.

\begin{enumerate}
    \item[a.] For a random initialisation of $x$, both the affine and one-hidden layer classifier cannot reconstruct any training images (see~\cref{fig:linear_classifier_1b} and \cref{fig:one_hidden_layer_1b}). 
The resulting images look like random noise. We additionally track the distance between ground truth parameters $\theta^*$ and the values of $\theta$ of the classifier after solving the reconstruction problem. For a random initialisation of $x$, as highlighted in \cref{tab:affine_theta} this distance is small for both the affine and one-hidden layer classifier. This shows that the optimisation problem converged even though the visual results for $x$ do not resemble any training data.

\item[b.] As shown in \cref{fig:linear_classifier_2c}, solving the bilevel optimisation problem when initialising $x$ with ground truth training images, the affine classifier recovers the training data perfectly. For the same initialisation, even the one-hidden layer classifier recovers images resembling the training data, as highlighted in \cref{fig:one_hidden_layer_2c}. In both cases, the distance between $\theta^*$ and $\theta$ remains small.

\item[c.] We further test initialising $x$ with a different partition of the dataset. Visual results for a random initialisation of $\theta$ for an affine and one-hidden layer classifier are depicted in \cref{fig:linear_classifier_3} and \cref{fig:one_hidden_classifier_3}, respectively. In contrast to a random initialisation of $x$, the recovered images for both classifiers are valid natural images. When comparing the reconstructions with the images used for initialisation, it is clear that the approach is not reconstructing the true training data but rather the same images that were used for the intialisation (see \cref{fig:linear_classifier_3b_CIFAR_init} and \cref{fig:one_hidden_layer_3b_CIFAR_init}). As in previous experiments, the distance between $\theta^*$  and $\theta$ remains small for both the affine and one-hidden layer classifier.

\end{enumerate}
\begin{figure*}[tb]
    \centering
    \includegraphics[trim=0cm 2cm 9.66cm 0cm, clip, width=0.60\textwidth]{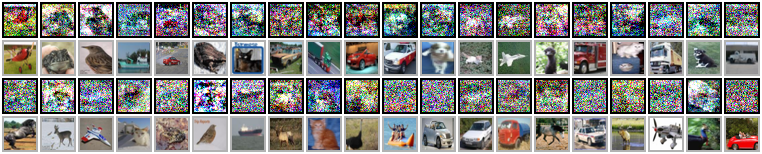}
    \caption{DecoReco reconstructions (black BB) for a random init. of $x$ without finetuning the variance of the distribution vs nearest neighbour of training set (gray BB).}
    \label{fig:decoreco_sf1}
\end{figure*}
In summary, all previous experiments show that the quality of the reconstructions highly depends on the initialisation of $x$.
With regard to privacy, the results signify that even when reconstructing natural images that could be part of the training dataset, an adversary yet cannot identify whether it actually belongs to the training set. As an initialisation with a different partition of images of the training dataset yields reconstructing even those images used for initialisation while maintaining a small distance between $\theta^*$ and $\theta$, there is no privacy breach.

\subsection{DecoReco formulation}
Having identified the importance of the initialisation of $x$ for the bilevel formulation, we explore the effect of the same initialisations on previous approaches  -- analysing if the results transfer to other formulations or are a peculiarity of our proposed approach.  
As to the best of our knowledge, only Buzaglo et al. \cite{buzaglo2023deconstructing} consider the problem of reconstructing training data with the same settings, we will focus on their approach, which,  for simplicity, in the following, we will refer to as DecoReco. We use a three linear layer neural network with bias terms. We do not modify the reconstruction setting used in \cite{buzaglo2023deconstructing} and only adapt the initialisation of $x$. 
Initialising the DecoReco approach randomly with $x \sim \mathcal{N}(0,1)$, i.e., without finetuning the variance of the distribution, \cref{fig:decoreco_sf1} shows that at most the first three reconstructions resemble natural images while still being very noisy.

\begin{figure}[t]
    \centering
    \includegraphics[width=0.5\linewidth]{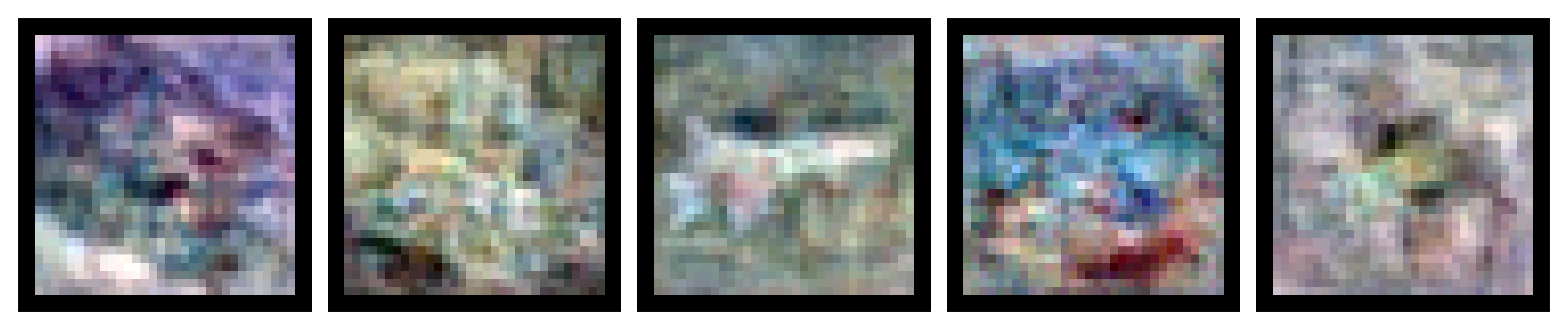}
    \caption{Four of these reconstructions are among the top 10 nearest neighbours to actual training images while one is not even in the top 50. Can you guess which one?}
    \label{fig:decoreco_quiz}
\end{figure}

\begin{figure}[b]
    \centering
    \includegraphics[width=0.5\linewidth]{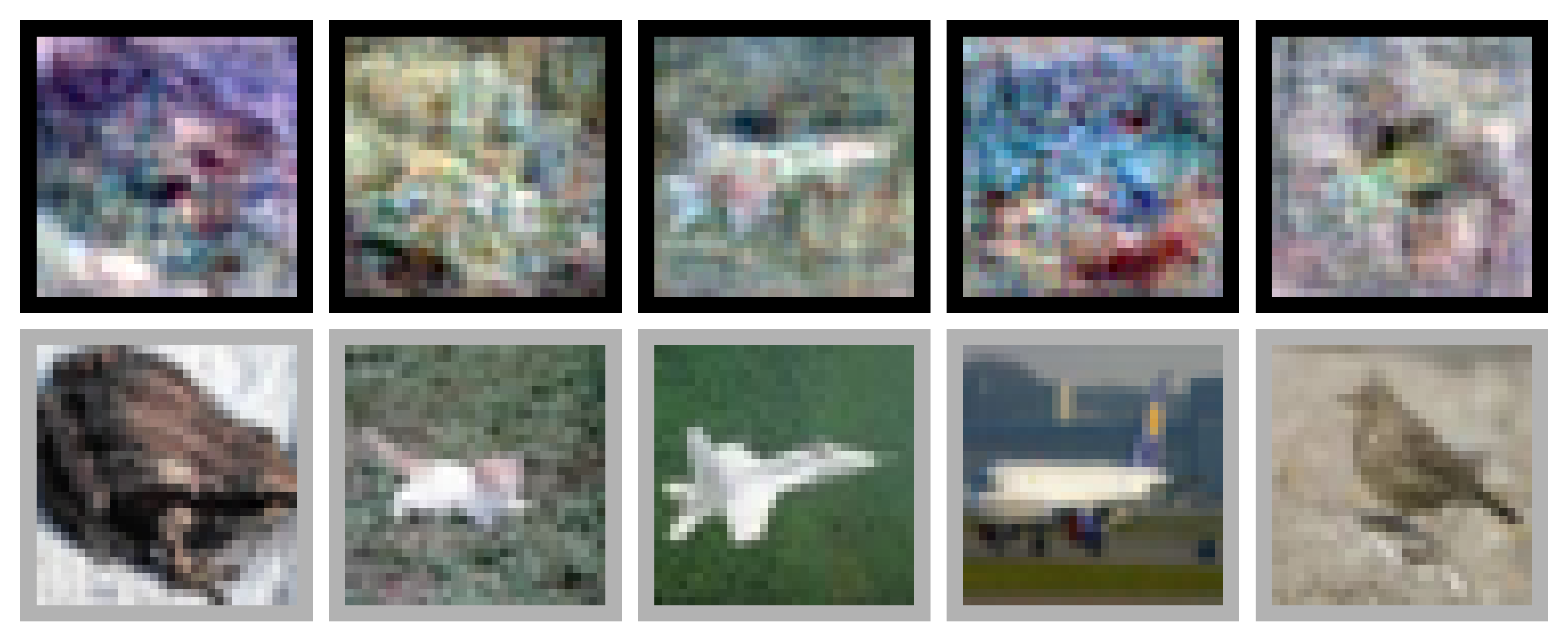}
    \caption{Solution: Clearly,  the 4\textsuperscript{th} reconstruction (top row) is not within the top 50 nearest neighbours to actual training images (bottom row).}
    \label{fig:decoreco_quiz_solution}
\end{figure}

In the following, we use the very specific scale factor used for the experiments in \cite{buzaglo2023deconstructing}, i.e., fine-tuning the variance. \cref{fig:decoreco_quiz} shows four reconstructions that are ranked within the top ten nearest neighbours within the actual training set. One image however is only ranked $51^{\text{st}}$. Only when
 looking at the reconstructions in combination with the training images, one is able to distinguish the four high  ranked images from the low-ranked (see \cref{fig:decoreco_quiz_solution}). The fourth image could be a realistic one comparable in quality to other reconstructions that resemble true training images more closely (images 1-3; 5). 
 
\noindent 

\begin{figure*}[t]
    \centering
    \includegraphics[trim=0cm 5.8cm 0cm 0cm, clip, width=0.80\textwidth]{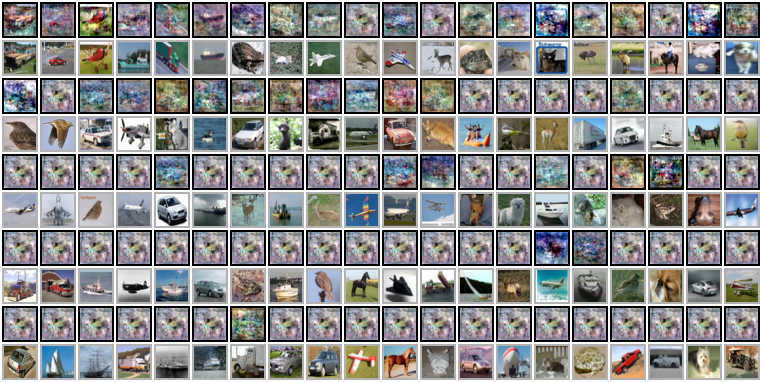}
    \caption{Reconstructions of the DecoReco approach (black BB) and their nearest neighbours withing the training dataset (gray BB) for a random init. of $x$. Lower ranked images without any resemblance to their nearest neighbour in some cases do resemble valid training set candidates.} 
    \label{fig:decoreco_rnd}
\end{figure*}

\setlength{\intextsep}{0pt}
\noindent%
An overview of the first $40$ reconstructions can be found in \cref{fig:decoreco_rnd}. This further underpins the findings for the bilevel optimisation formulation and highlights, that without having access to the training data, an adversary cannot identify whether a reconstructed natural image is part of the training dataset.

\begin{figure*}[t]
    \centering
    \begin{subfigure}[b]{0.48\textwidth}
        \includegraphics[trim=0cm 2cm 9.66cm 0cm, clip,width=\textwidth]{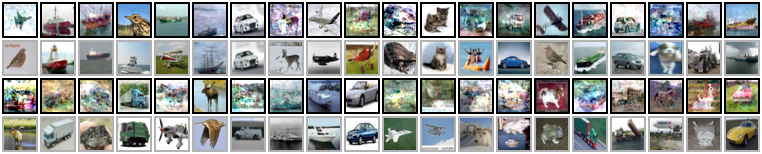}
        \caption{DecoReco reconstructions (black BB) vs their nearest neighbour of the training dataset (gray BB).}
        \label{fig:decoreco_cifar_gt_comparison}
    \end{subfigure}
    \hfill
    \begin{subfigure}[b]{0.48\textwidth}
            \includegraphics[trim=0cm 2cm 9.66cm 0cm, clip,width=\textwidth]{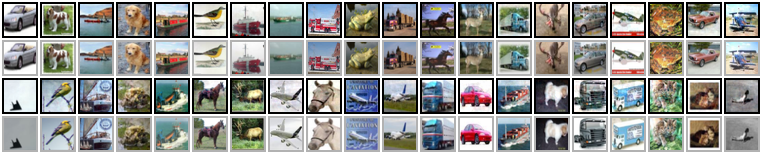}
        \caption{DecoReco reconstructions (black BB) vs their nearest neighbour of the data partition used for initialisation (gray BB).}
        \label{fig:decoreco_cifar_cifar_comparison}
    \end{subfigure}
    \caption{Reconstructions of DecoReco approach when initialising $x$ with CIFAR images of a distinct partition. Reconstructions resemble the images of the partition used for intialisation rather than the ground truth training images.}
    \label{fig:decoreco_cifar}
\end{figure*}

We further examine the effect of initialising the DecoReco approach with a distinct  partition of the CIFAR dataset, not overlapping with the partition used for training the classifier. 
For this initialisation, we obtain similar results as the bilevel problem with the same partition. 
\cref{fig:decoreco_cifar} provides an overview of reconstructions and their structurally closest counterparts within the ground truth training set (see \cref{fig:decoreco_cifar_gt_comparison}) and within the CIFAR image partition used for initialisation (see \cref{fig:decoreco_cifar_cifar_comparison}). Initialising the reconstruction 
problem with previously unseen images from the same distribution, the majority of the resulting images reflects the initialisation rather than the real training samples.
Interestingly, a single training image can be reconstructed (see \cref{fig:decoreco_cifar_gt_comparison} first row, 7\textsuperscript{th} sample from the left). 

To summarise, testing different initialisations on another formulation of the reconstruction problem confirms the findings in \cref{subsec:bilevel_optimisation_problem} and attests that those effects are not just peculiarities of the bilevel approach.
\subsection{Exploring the energy landscape of the optimisation problem}\label{sec:energy_landscape}
\begin{figure}[t]
    \centering
    \begin{minipage}{0.45\textwidth}
    \centering
    \begin{subfigure}[b]{0.48\textwidth}
    \centering
            \includegraphics[width=\textwidth,trim={0cm 0cm 0cm 0cm},clip]{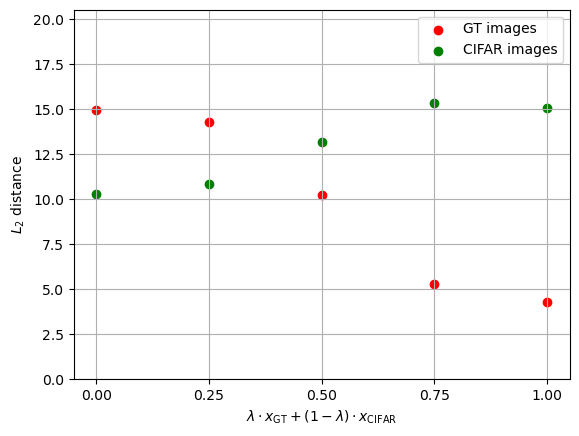}
        \caption{Bilevel.}
        \label{fig:grid_experiment_bilevel_2d_gt_cifar}
    \end{subfigure}
    \begin{subfigure}[b]{0.48\textwidth}
    \centering
            \includegraphics[width=\textwidth,trim={0cm 0cm 0cm 0cm},clip]{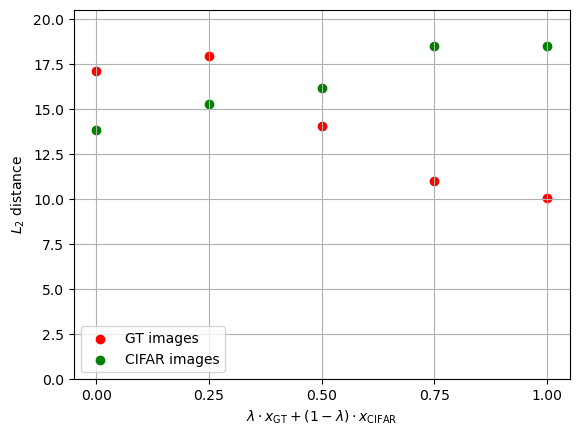}
        \caption{DecoReco.}
        \label{fig:grid_experiment_decoreco_2d_gt_cifar}
    \end{subfigure}
    \caption{Reconstruction results for interpolation between images of a distinct partition of the dataset and GT training samples as init. of $x$.}
    \label{fig:grid_experiment_2d}
    \end{minipage}
    \hfill
    \begin{minipage}{0.45\textwidth}
     \centering
    \begin{subfigure}[b]{0.48\textwidth}
        \includegraphics[width=\textwidth,trim={0.5cm 0cm 1.5cm 0cm},clip]{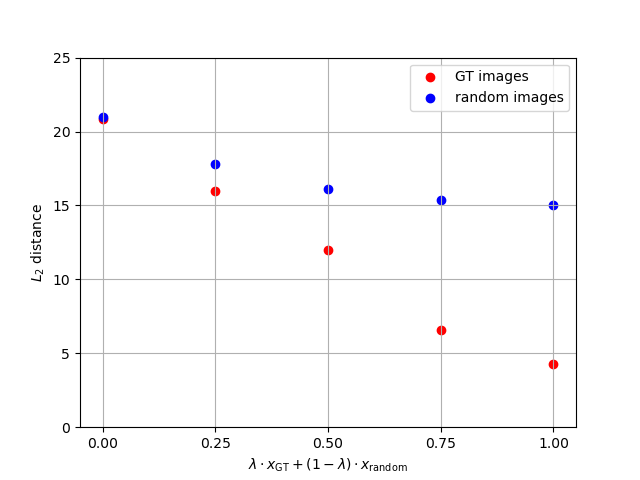}
        \caption{Bilevel.}
        \label{fig:grid_experiment_bilevel_2d_gt_random}
    \end{subfigure}
        \begin{subfigure}[b]{0.48\textwidth}
        \includegraphics[width=\textwidth]{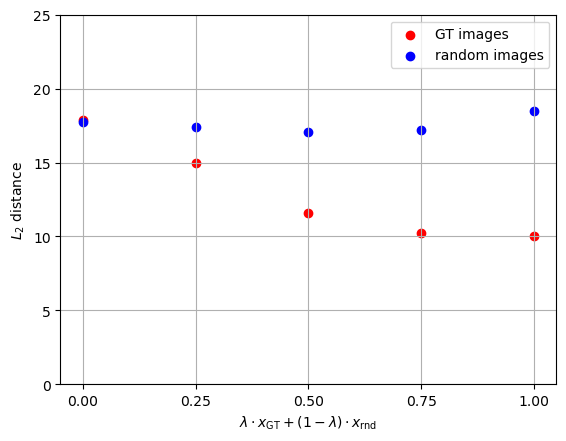}
        \caption{DecoReco.}
        \label{fig:grid_experiment_decoreco_2d_gt_random}
    \end{subfigure}
    \caption{Reconstruction results for interpolation between ground truth training samples and random noise images as init. of $x$.}
    \label{fig:grid_experiment_decoreco}        
    \end{minipage}
\end{figure}

Numerous local minima leading to a sensitivity to initialisations indicate a highly complex energy landscape of the optimisation problem. In the following, we empirically explore the landscape of both formulations, bilevel optimisation and  DecoReco. We do this by mixing the three different 
\begin{figure}[tb]
      \centering
    \begin{subfigure}[b]{0.24\textwidth}
        \includegraphics[width=\textwidth,trim={0cm 0cm 0cm 0cm},clip]{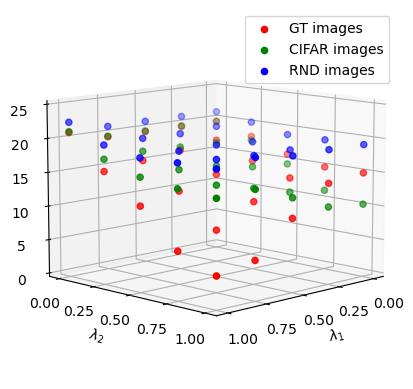}
        \caption{Bilevel.}
        \label{fig:grid_experiment_bilevel_3d}
    \end{subfigure}
    \begin{subfigure}[b]{0.24\textwidth}
            \includegraphics[width=\textwidth,trim={0cm 0cm 0cm 0cm},clip]{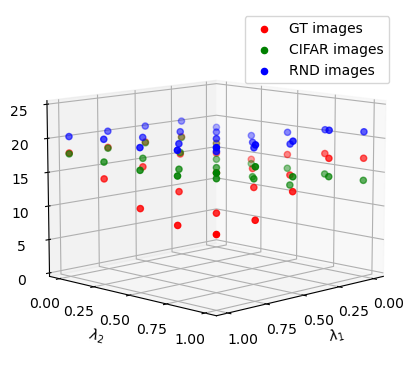}
        \caption{DecoReco.}
        \label{fig:grid_experiment_decoreco_3d}
    \end{subfigure}
    \caption{$L_2$ distance for weighting init. of $x$ from the ground truth (GT) training images, a distinct partition of the dataset (CIFAR) and random noise (RND).}
    \label{fig:grid_experiment_3d}
\end{figure} 
\noindent
 initialisation schemes used in previous experiments to generate new initialisations for $x$. In particular,
\begin{equation}
    x_{\text{init}} = \lambda_2 \Big( \lambda_1 \cdot x_{\text{GT}} + (1-\lambda_1)\cdot x_{\text{part.}} \Big) + (1-\lambda_2) \cdot x_{\text{rnd}}
\end{equation}
for constants $\lambda_1 \text{ and } \lambda_2$, $x_{\text{GT}}$ sampled from the training dataset, $x_{\text{part.}}$ sampled from the distinct partition of the CIFAR dataset and $x_{\text{rnd}} \sim \mathcal{U}(0, 1)$ to resemble random noise. We choose both $\lambda_1, \lambda_2 \in [0, 0.25, 0.5, 0.75, 1]$ to generate 25 initialisations for $x$.
The average $L_2$ distance between every reconstructed image and its structurally most similar ground truth image, image from the distinct CIFAR partition and random noise image, respectively, for both the bilevel optimisation formulation using an affine classifier and the DecoReco approach are highlighted in \cref{fig:grid_experiment_3d}. 
While the bilevel formulation on average performs better in recovering ground truth images if $x_{\text{init}}$ is close enough to the ground truth (see \cref{fig:grid_experiment_bilevel_3d}), the average distance increases quicker for a higher proportion of $x_{\text{rnd}}$ than in the DecoReco formulation (see \cref{fig:grid_experiment_decoreco_3d}). 
These results are further highlighted in \cref{fig:grid_experiment_2d} where we fix $\lambda_2 = 1$.
\cref{fig:grid_experiment_bilevel_2d_gt_cifar} 
and \cref{fig:grid_experiment_decoreco_2d_gt_cifar} highlight the average $L_2$ distance when interpolating between ground truth images and CIFAR images of a different partition of the dataset.
For $\lambda < 0.5$, the average $L_2$ distance between the reconstructions and images of the distinct CIFAR partition (green) is smaller than the distance to the ground truth training images for both approaches. This highlights that both approaches if initialised sufficiently close to images of the dataset, reconstruct the initialisations rather than actual training images. In case of the bilevel formulation, the average $L_2$ distance becomes much smaller than for the DecoReco approach. 
Similarly, for $\lambda_1 =1$, the initialisation of $x$ interpolates between ground truth training images and random noise. \cref{fig:grid_experiment_bilevel_2d_gt_random} and  \cref{fig:grid_experiment_decoreco_2d_gt_random} show the average $L_2$ distance of every reconstruction and its nearest neighbour within the training dataset (red) and a set of random noise images (blue). Increasing $\lambda$ increases the influence of ground truth training images on the initialisation, which for both formulations of the optimisation problem, results in reconstructions converging towards ground truth training images. In case of the bilevel formulation, the average $L_2$ distance becomes much smaller than for the DecoReco approach.
We conjecture that the energy landscape of the DecoReco formulation could be a smoothed version with fewer local minima than that of the actual bilevel formulation.

\section{Conclusion}
In this work, we formulated the problem of reconstructing training data from neural network parameters as a bilevel optimisation problem. We showed that the reconstructions highly depend on the initialisation of the input $x$, for both the bilevel and the DecoReco formulation introduced in \cite{buzaglo2023deconstructing}. For an intialisation sufficiently close to $x$, it is possible to reconstruct training data. 
It is however also possible to reconstruct other natural images that resemble potential training samples without them being part of the actual training dataset.
Consequently, with regards to privacy, the results signify that even when reconstructing natural images that could be part of the training dataset, an adversary yet cannot identify whether it is a valid or invalid training data sample with certainty.

\section*{Acknowledgements}
C.R. acknowledges support from the Cantab Capital Institute for the Mathematics of Information (CCIMI) and the EPSRC grant EP/W524141/1. K.V.G. and M.M. acknowledge the support of the Lamarr Institute for Machine Learning and Artificial Intelligence. C.B.S. acknowledges support from the Philip Leverhulme Prize, the Royal Society Wolfson Fellowship, the EPSRC advanced career fellowship EP/V029428/1, EPSRC grants EP/S026045/1 and EP/T003553/1, EP/N014588/1, EP/T017961/1, the Wellcome Innovator Awards 215733/Z/19/Z and 221633/Z/20/Z, the European Union Horizon 2020 research and innovation programme under the Marie Skodowska-Curie grant agreement No. 777826 NoMADS, the Cantab Capital Institute for the Mathematics of Information and the Alan Turing Institute. 


\begin{thebibliography}{29}
\bibitem{balle2022reconstructing}
Balle B.,  Cherubin H., and  Hayes J.
\newblock Reconstructing training data with informed adversaries.
\newblock In \emph{IEEE Symposium on Security and Privacy (SP)},
2022.

\bibitem{buzaglo2023deconstructing}
Buzaglo G., Haim N., Yehudai G., Vardi G., Oz  Y., Nikankin Y., et~al.
\newblock Deconstructing data reconstruction: Multiclass, weight decay and general losses.
\newblock In \emph{Thirty-seventh Conference on Neural Information Processing Systems}, 2023.


\bibitem{carlini2023quantifying}
Carlini N.,  Ippolito D.,  Jagielski M.,  Lee K.,  Tramer F., and  Zhang C.
\newblock Quantifying memorization across neural language models.
\newblock In \emph{International Conference on Learning Representations}, 2023.

\bibitem{fredrikson2015model}
 Fredrikson M., Jha S., and  Ristenpart T.
\newblock Model inversion attacks that exploit confidence information and basic countermeasures
\newblock In \emph{Proc. 22nd ACM SIGSAC conference on computer and communications security}, 2015.

\bibitem{Geiping_2019_ICCV}
 Geiping J. and  Moeller M.
\newblock Parametric majorization for data-driven energy minimization methods.
\newblock In \emph{International Conference on Computer Vision}, 2019.

\bibitem{geiping2020inverting}
 Geiping J.,  Bauermeister H.,  Dr{\"o}ge H., and  Moeller M.
\newblock Inverting gradients-how easy is it to break privacy in federated learning?
\newblock \emph{Advances in Neural Information Processing Systems}, 33:\penalty0 16937--16947, 2020.

\bibitem{guo2022bounding}
 Guo C.,  Karrer B.,  Chaudhuri K., and  Maaten L.
\newblock Bounding training data reconstruction in private (deep) learning.
\newblock In \emph{International Conference on Machine Learning},  2022.

\bibitem{haim2022reconstructing}
 Haim N.,  Vardi G., Yehudai G., Shamir O., and  Irani M.
\newblock Reconstructing training data from trained neural networks.
\newblock In \emph{Advances in Neural Information Processing Systems}, 35:\penalty0 22911--22924, 2022.

\bibitem{hatamizadeh2022gradvit}
Hatamizadeh A., Yin H., Roth H.~R.,Li W. , Kautz J., Xu D., and Molchanov P.
\newblock Gradvit: Gradient inversion of vision transformers.
\newblock In \emph{Proc. IEEE/CVF Conference on Computer Vision and Pattern Recognition},  2022.

\bibitem{hayes2023bounding}
Hayes J., Mahloujifar S., and Balle B.
\newblock Bounding training data reconstruction in dp-sgd.
\newblock \emph{Advances in neural information processing systems}, 2023.

\bibitem{jeon2021gradient}
 Jeon J., Lee K., Oh S., and Ok J.
\newblock Gradient inversion with generative image prior.
\newblock \emph{Advances in neural information processing systems}, 2021.

\bibitem{ji2020}
 Ji Z. and Telgarsky  M.
\newblock Directional convergence and alignment in deep learning.
\newblock In \emph{Advances in Neural Information Processing Systems}, 2020.

\bibitem{krizhevsky2009learning}
 Krizhevsky A. 
 \newblock Learning multiple layers of features from tiny images.
\newblock 2009.

\bibitem{loo2023dataset}
 Loo N., Hasani R., Lechner M., and Rus D.
\newblock Understanding Reconstruction Attacks with the Neural Tangent Kernel and Dataset Distillation.
\newblock In \emph{International Conference on Learning Representations}, 2024.

\bibitem{lorraine2020optim}
Lorraine J., Vicol P., and Duvenaud D.
\newblock Optimizing Millions of Hyperparameters by Implicit Differentiation.
\newblock In \emph{Proc. Twenty Third International Conference on Artificial Intelligence and Statistics}, 2020.

\bibitem{Lyu2020Gradient}
 Lyu K. and Li J.
\newblock Gradient descent maximizes the margin of homogeneous neural networks.
\newblock In \emph{International Conference on Learning Representations}, 2020.

\bibitem{nasr2019comprehensive}
 Nasr M., Shokri R., and Houmansadr A.
\newblock Comprehensive privacy analysis of deep learning: Passive and active white-box inference attacks against centralized and federated learning.
\newblock In \emph{IEEE symposium on security and privacy (SP)}, 
2019.

\bibitem{phong2017privacy}
 Phong L. T., Aono Y., Hayashi T., Wang L., and Moriai S. 
\newblock Privacy-preserving deep learning: Revisited and enhanced.
\newblock In \emph{Proc. 8th International Conference on Applications and Techniques in Information Security}, pp. 100--110. Springer, 2017.

\bibitem{shokri2017membership}
Shokri R., Stronati M., Song C., and Shmatikov V.
\newblock Membership inference attacks against machine learning models.
\newblock In \emph{IEEE symposium on security and privacy}, pp. 3--18, 2017.

\bibitem{somepalli2023diffusion}
 Somepalli G., Singla V., Goldblum M., Geiping J., and Goldstein T.
\newblock Diffusion art or digital forgery? investigating data replication in diffusion models.
\newblock In \emph{Proc. IEEE/CVF Conference on Computer Vision and Pattern Recognition}, 
2023.

\bibitem{song2013stochastic}
 Song S., Chaudhuri  K., and Sarwate A.~D.
\newblock Stochastic gradient descent with differentially private updates.
\newblock In \emph{IEEE global conference on signal and information processing}, pp. 245--248, 2013.

\bibitem{wang2018dataset}
Wang T., Zhu J. Y., Torralba A., and Efros A.
\newblock Dataset distillation.
\newblock \emph{arXiv preprint arXiv:1811.10959}, 2018.

\bibitem{wang2021variational}
Wang, K. C., Fu, Y., Li, K., Khisti, A., Zemel, R., and Makhzani, A.
\newblock Variational model inversion attacks.
\newblock In \emph{Neural Information Processing Systems},
2021.

\bibitem{ye2023initialisation}
 Ye J., Zhu Z., Liu F., Shokri  R., and Cevher V.
\newblock initialisation matters: Privacy-utility analysis of overparameterized neural networks.
\newblock In \emph{Thirty-seventh Conference on Neural Information Processing Systems}, 2023.

\bibitem{yin2021see}
 Yin H., Mallya A., Vahdat A., Alvarez J.~M., Kautz J., and Molchanov P.
\newblock See through gradients: Image batch recovery via gradinversion.
\newblock In \emph{Proc. IEEE/CVF Conference on Computer Vision and Pattern Recognition}, pp. 16337--16346, 2021.

\bibitem{zhang2023understanding}
 Zhang H., Hong J., Deng Y., Mahdavi M., and Zhou J.
\newblock Understanding deep gradient leakage via inversion influence functions.
\newblock In \emph{Thirty-seventh Conference on Neural Information Processing Systems}, 2023.

\bibitem{zhao2020idlg}
 Zhao B., Mopuri K.~R., and Bilen H.
\newblock idlg: Improved deep leakage from gradients.
\newblock \emph{arXiv preprint arXiv:2001.02610}, 2020.

\bibitem{zhu2020r}
Zhu J.  and Blaschko M.~B.
\newblock R-gap: Recursive gradient attack on privacy.
\newblock In \emph{International Conference on Learning Representations}, 2020.

\bibitem{zhu2019deep}
 Zhu L., Liu Z., and Han S.
\newblock Deep leakage from gradients.
\newblock In \emph{Advances in neural information processing systems}, 32, 2019.

\end{thebibliography}
\end{document}